\documentclass{article} 
\usepackage{iclr2025_conference,times}

\usepackage{hyperref}
\usepackage{url}

\definecolor{mycitecolor}{RGB}{0, 0, 128}
\definecolor{myrefcolor}{RGB}{199, 29, 29}

\usepackage{amsmath}
\usepackage{amssymb}
\usepackage{graphicx}
\usepackage{multirow}
\usepackage{makecell}
\usepackage{booktabs}
\usepackage{wrapfig}
\usepackage{tabularx}
\usepackage{algorithm}
\usepackage{algorithmic}
\usepackage[switch]{lineno}
\usepackage{xcolor}
\usepackage{dsfont}
\hypersetup{colorlinks, linkcolor=myrefcolor, citecolor=mycitecolor, urlcolor=mycitecolor}

\title{\CoDiCast: Conditional Diffusion Model for Global Weather Prediction with Uncertainty Quantification}

\iclrfinalcopy 
\author{
Jimeng Shi$^{1}$, 
Bowen Jin$^{2}$, 
Jiawei Han$^{2}$, 
Sundararaman Gopalakrishnan$^{3}$,
Giri Narasimhan$^{1}$ \\
$^1$Florida International University, 
\\$^2$University of Illinois Urbana-Champaign, \\
$^3$Hurricane Research Division, National Oceanic and Atmospheric Administration \\
\texttt{\{jshi008,giri\}@fiu.edu, \{bowenj4,hanj\}@illinois.edu}\\
\texttt{sundararaman.g.gopalakrishnan@noaa.gov}
}

\newcommand{\CoDiCast}{\texttt{CoDiCast}}

\begin{document}

\maketitle

\begin{abstract}
Accurate weather forecasting is critical for science and society. However, existing methods have not achieved the combination of high accuracy, low uncertainty, and high computational efficiency simultaneously.
On one hand, traditional numerical weather prediction (NWP) models are computationally intensive because of their complexity.
On the other hand, most machine learning-based weather prediction (MLWP) approaches offer efficiency and accuracy but remain deterministic, lacking the ability to capture forecast uncertainty.
To tackle these challenges, we propose a conditional diffusion model, \CoDiCast, to generate global weather prediction, integrating accuracy and uncertainty quantification at a modest computational cost. 
The key idea behind the prediction task is to generate realistic weather scenarios at a \emph{future} time point, conditioned on observations from the \emph{recent past}. 
Due to the probabilistic nature of diffusion models, they can be properly applied to capture the uncertainty of weather predictions. Therefore, we accomplish uncertainty quantifications by repeatedly sampling from stochastic Gaussian noise for each initial weather state and running the denoising process multiple times. 
Experimental results demonstrate that \CoDiCast~outperforms several existing MLWP methods in accuracy, and is faster than NWP models in the inference speed. 
Our model can generate 6-day global weather forecasts, at 6-hour steps and $5.625^\circ$ latitude-longitude resolutions, for over 5 variables, in about 12 minutes on a commodity A100 GPU machine with 80GB memory. 
Our code is available at \url{https://github.com/JimengShi/CoDiCast}.
\end{abstract}

\section{Introduction}
\label{sec:intro}
Weather prediction describes how the weather states evolve by mapping the current weather states to future weather states \citep{palmer2012towards}.
Accurate weather forecasting is crucial for a wide range of societal activities, from daily planning to disaster preparedness \citep{merz2020impact, shi2024fidlar}. 
For example, governments, organizations, and individuals rely heavily on weather forecasts to make informed decisions that can significantly impact safety, economic efficiency, and overall well-being. 
However, weather predictions are intrinsically uncertain largely due to the complex and chaotic nature of atmospheric processes \citep{slingo2011uncertainty}. 
Therefore, assessing the range of probable weather scenarios is significant, enabling informed decision-making.

Traditional numerical weather prediction (NWP) methods achieve weather forecasting by approximately solving the differential equations representing the integrated system between the atmosphere, land, and ocean \citep{price2023gencast, nguyen2023climax}.
However, running such an NWP model can produce only one possibility of the forecast, which ignores the weather uncertainty.
To solve this problem, \emph{Ensemble forecast}\footnote{Generating a set of forecasts, each of which represents a single possible scenario.} of multiple models is often employed to model the probability distribution of different future weather scenarios \citep{palmer2019ecmwf, leinonen2023latent}.
While such NWP-based ensemble forecasts effectively model the weather uncertainty, physics-based models inherently make restrictive assumptions of atmospheric dynamics, and running multiple NWP models requires extreme computational costs \citep{rodwell2007using}.

In recent years, machine learning (ML)-based weather predictions (MLWP) have been proposed to challenge NWP-based prediction methods \citep{ben2024rise, bulte2024uncertainty}. 
They have achieved enormous success with comparable accuracy and a much (usually three orders of magnitude) lower computational overhead. 
They are typically trained to learn weather patterns from a huge amount of historical data and predict the mean of the probable trajectories by minimizing the mean squared error (MSE) of model forecasts \citep{hewage2021deep}. 
Representative work includes Pangu \citep{bi2023accurate}, GraphCast \citep{lam2023learning}, ClimaX \citep{nguyen2023climax}, ForeCastNet \citep{pathak2022fourcastnet}, Fuxi \citep{chen2023fuxi}, Fengwu and \citep{chen2023fengwu}.
Despite the notable achievements of these MLWP methods, most of them are deterministic \citep{kochkov2024neural}, falling short in capturing the uncertainty in weather forecasts \citep{jaseena2022deterministic}.

To compute the uncertainty for ML models, two methods exist. 
Perturbing initial conditions \citep{morley2018perturbed} helps estimate the aleatoric uncertainty (data noise), while the Monte Carlo Dropout approach \citep{gal2016dropout} estimates epistemic uncertainty (model uncertainty) \citep{siddique2022survey}.
However, neither approach fully captures uncertainty in both the input conditions and the evolution of weather models.
Additionally, these methods require manual tuning of perturbations and dropout rates, which can negatively impact model accuracy. 
These limitations motivate us to explore an approach for comprehensive uncertainty quantification without the loss of accuracy.

Denoising probabilistic diffusion models (DDPMs) \citep{ho2020denoising} stand out as a probabilistic type of generative models, which can generate high-quality images.
By explicitly and iteratively modeling the noise additive and its removal, DDPMs can capture intricate details and textures of images.
Furthermore, controllable diffusion models \citep{rombach2022high,zhang2023adding} enable the generation process to be guided by specific attributes or conditions, e.g., class labels, textual descriptions, or other auxiliary information. By doing so, the models can generate images that adhere to the specified conditions.
This inspires us to consider the weather ``prediction'' tasks as ``generation'' tasks - generating plausible weather scenarios with conditional diffusion models. Promising potentials could be the following: 
(1) Weather numerical data is usually a 2-D grid over latitude and longitude, sharing a similar modality with the image. Diffusion models can capture the intricate weather distribution with iterative denoising.
(2) Weather states from the recent past (i.e., initial conditions) can be injected into diffusion models to guide the generation of future weather evolution.
(3) More notably, the starting noise sampling from the Gaussian distribution can mimic the aleatoric uncertainty while iteratively adding and removing noise captures the epistemic uncertainty.
These features prompt probabilistic diffusion models to generate a set of diverse weather scenarios rather than a single deterministic one. This capability makes them well-suited for modeling the uncertain nature of weather evolution.
Our contributions are presented as follows:
\begin{itemize}
    \item We identify the shortcomings of current weather prediction methods. NWP-based methods are limited to restrictive assumptions and computationally intensive. Moreover, a single deterministic NWP- and MLWP-based method cannot achieve uncertainty quantification.
    \item To address these problems, we propose \CoDiCast, a conditional diffusion model for global weather prediction conditioning on observations from the recent past while probabilistically modeling the uncertainty. In addition, we use the cross-attention mechanism to effectively integrate conditions into the denoising process to guide the generation tasks.
    \item We conduct extensive experiments on a decade of ERA5 reanalysis data from the European Centre for Medium-Range Weather Forecasts (ECMWF), and evaluate our method against several state-of-the-art models in terms of accuracy,  efficiency, and uncertainty. It turns out that \CoDiCast~achieves an essential trade-off among these valuable properties.
\end{itemize}
\section{Preliminaries}
\label{sec:prel}
In this section, we introduce the problem formulation of global weather prediction and briefly review Denoising Diffusion Probabilistic Models (DDPMs) \citep{ho2020denoising}.

\subsection{Problem Formulation}
\label{sec:problem}
%
%
\textbf{Deterministic Global Weather Predictions.} 
Given the input consisting of the weather state(s), $X^t \in \mathbb{R}^{H \times W \times C}$ at time $t$, the problem is to predict a point-valued weather state, $X^{t+\Delta t} \in \mathbb{R}^{H \times W \times C}$ at a future time point $t+\Delta t$.
$H \times W$ refers to the spatial resolution of data which depends on how densely we grid the globe over latitudes and longitudes, $C$ refers to the number of channels (i.e., weather variables), and the superscripts $t$ and $t+\Delta t$ refer to the current and future time points. The long-range multiple-step forecasts could be achieved by autoregressive modeling or direct predictions. 

\begin{wrapfigure}{r}{0.42\textwidth}
\vskip -0.4cm
\centering
    \includegraphics[width=0.42\textwidth]{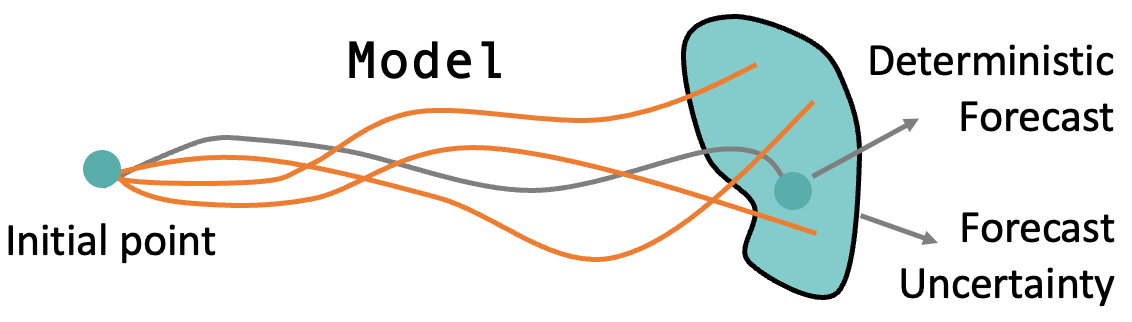}
\vspace{-4mm}
\caption{Deterministic vs Probabilistic.}
\vskip -0.6cm
\label{fig:prob_prediction}
\end{wrapfigure}
\noindent \textbf{Probabilistic Global Weather Predictions.} Unlike the deterministic models that output point-valued predictions, probabilistic methods model the probability of future weather state(s) as a distribution $P(X^{t+\Delta t} \mid X^t)$, conditioned on the state(s) from the recent past. Probabilistic predictions are appropriate for quantifying the forecast uncertainty and making informed decisions.

\subsection{Denoising Diffusion Probabilistic Models}
\label{sec:ddpm}
A denoising diffusion probabilistic model (DDPM) \citep{ho2020denoising} generates target samples by learning a distribution $p_\theta(x_0)$ that approximates the target distribution $q(x_0)$.
DDPM comprises a \emph{forward diffusion} process and a \emph{reverse denoising} process. 
The \emph{forward process} 
transforms an input $x_0$ with a data distribution of $q(x_0)$ to a Gaussian noise vector $x_N$ in $N$ diffusion steps. It can be described as a Markov chain that gradually adds Gaussian noise to the input according to a variance schedule $\{\beta_1, \dots, \beta_N\}$:
\begin{equation}
    q(x_n \mid x_{n-1}) = \mathcal{N}\left(x_n; \sqrt{1 - \beta_n} x_{n-1}, \beta_n \textbf{I}\right), \mathrm{ and}
\end{equation}
\begin{equation}
    q(x_{1:N} \mid x_0) = \prod_{n=1}^{N} q(x_n \mid x_{n-1}), n \in [1, N],
\end{equation}
where at each step $n$, the diffused sample $x_n$ is obtained from $x_{n-1}$ as described above. 
Multiple steps of the forward process can be described as follows in a closed form:
\begin{equation*}
    q(x_n \mid x_0) = \mathcal{N}\left(x_n; \sqrt{\bar{\alpha}_n}x_0, (1-\bar{\alpha}_n) \textbf{I}\right), 
    \label{eq:forward}
\end{equation*}
where $\alpha_n = 1 - \beta_n$ and $\bar{\alpha}_n = \prod_{s=1}^{n} \alpha_s$. Thus, $x_n = \sqrt{\bar{\alpha}_n} x_0 + \sqrt{1-\bar{\alpha}_n} \epsilon$, with $\epsilon$ sampled from $\mathcal{N}(\textbf{0}, \textbf{I})$.

In the \emph{reverse process}, the \emph{denoiser} network is used to recover $x_0$ by stepwise denoising starting from the pure noise sample, $x_N$. This process is formally defined as:
\begin{equation}
    p_\theta(x_{0:N}) = p(x_N)\prod_{n=1}^{N} p_\theta(x_{n-1} \mid x_{n}),
    \label{eq:reverse}
\end{equation}
where $p_\theta(x_n)$ is the distribution at step $n$ parameterized by $\theta$.

For each iteration, $n \in [1, N]$, diffusion models are trained to minimize the following KL-divergence:
\begin{equation}
    \mathcal{L}_n = D_{KL}\left(q(x_{n-1} \mid x_n) \mid\mid p_\theta(x_{n-1} \mid x_n) \right).
    \label{eq:kl_loss}
\end{equation}
where $q(x_{n-1} | x_n)$ can be computed from $q(x_n | x_{n-1})$ using Bayes rule and the multistep forward process equation above.

\section{Methodology}
\label{sec:method}
\begin{figure*}[ht!]
    \centering
    \includegraphics[width=0.85\textwidth]{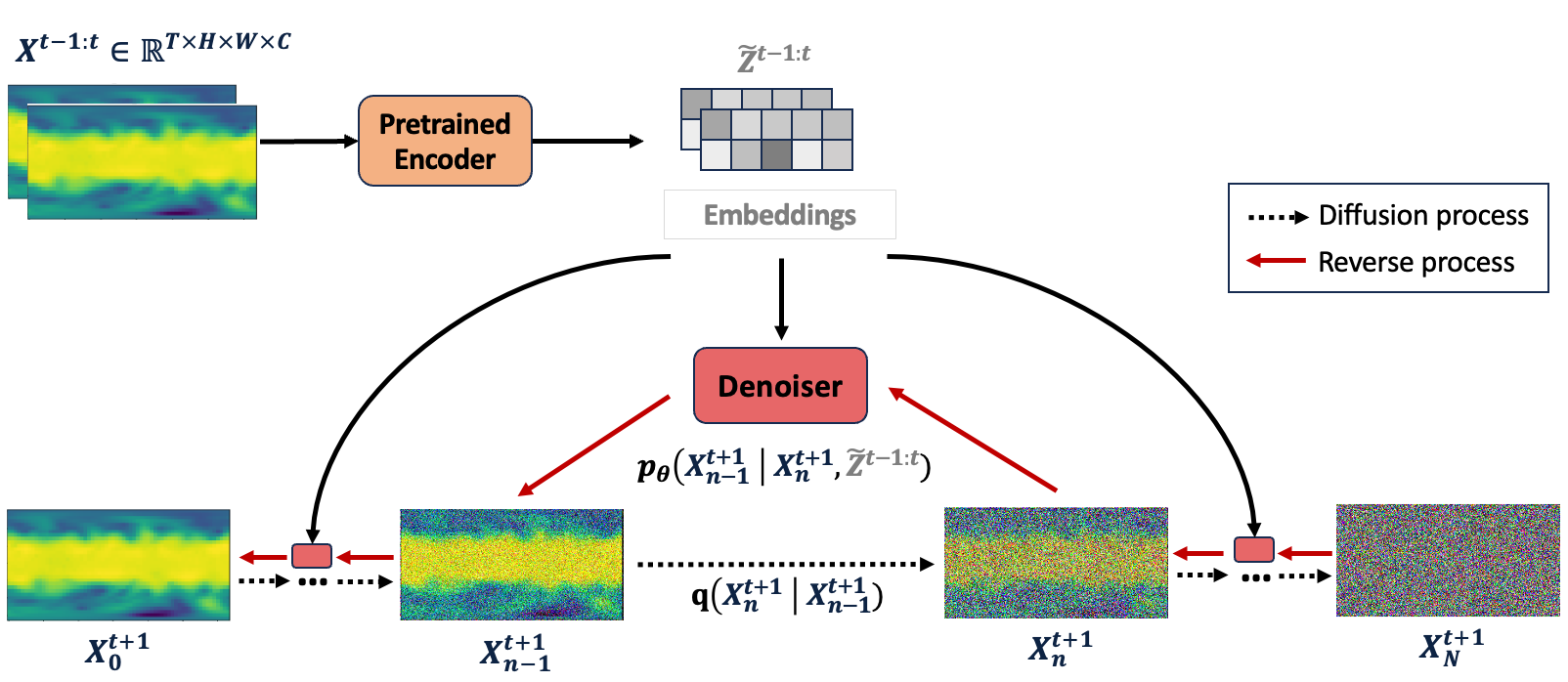}
    \caption{Framework of \CoDiCast~for global weather forecast. The superscript $T$ and the subscript $N$ denote the time point and iteration step of adding/denoising noise. $H$ and $W$ represent the height (\#latitude) and width (\#longitude) of grid data. $C$ is the number of variables of interest.}
    \label{fig:framework}
\end{figure*}
This section introduces our approach for global weather prediction, \texttt{CoDiCast}, implemented as a conditional diffusion model.
The key idea is to consider ``prediction'' tasks as ``generation'' tasks while conditioning on the context guidance of past observation(s). 
An overview of the proposed \texttt{CoDiCast} is shown in Figure \ref{fig:framework}.

\subsection{Forward Diffusion Process}
The forward diffusion process is straightforward. Assuming the current time point is $t$, for the sample at time point $t+1$, $X^{t+1}_0 \in \mathbb{R}^{H \times W \times C}$, which is of interest to predict, we first compute the diffused sample by gradually adding noise until the $N^{th}$ iteration (see the dotted lines in Figure \ref{fig:framework}):
\begin{equation}
    X^{t+1}_n = \sqrt{\bar{\alpha}_n} \cdot X^{t+1}_0 + \sqrt{1-\bar{\alpha}_n} \epsilon,
\end{equation}
where $\epsilon$ is sampled from $\mathcal{N}(\mathbf{0}, \mathbf{I})$ with the same size as $X^{t+1}_{0}$, and $\bar{\alpha}$ is same as that in Eq. (\ref{eq:forward}).

\subsection{Reverse Conditional Denoising Process}
\CoDiCast~models the probability distribution of the future weather state conditioning on the current and previous weather states. More specifically, we exploit a pre-trained encoder to learn conditions as embedding representations of the past observations $X^{t-1}$ and $X^t$, which are used to control and guide the synthesis process. 
Compared to modeling the past observations in the original space, we found that our embedding representations in the latent space work better.
\begin{equation}
    p_\theta(X_{0:N}^{t+1} \mid \tilde{Z}^{t-1:t}) = p(X_N^{t+1})\prod_{n=1}^{N} p_\theta(X_{n-1}^{t+1} \mid X_{n}^{t+1}, \tilde{Z}^{t-1:t}),
    \label{eq:condition_reverse}
\end{equation}
where $X_N^{t+1} \sim \mathcal{N}(\textbf{0}, \textbf{I})$, $\tilde{Z}^{t-1:t}$ is the embedding representation as shown in Eq. (\ref{eq:encoder}).

After prediction at the first time point is obtained, a forecast trajectory, $X^{1:T}$, of length $T$, can be auto-regressively modeled by conditioning on the predicted ``previous'' states.
\begin{equation}
    p_\theta(X_{0:N}^{1:T}) = \prod_{t=1}^{T} p(X_N^{t}) \prod_{n=1}^{N} p_\theta(X_{n-1}^{t} \mid X_{n}^{t}, \tilde{Z}^{t-2:t-1}).
    \label{eq:multi_prediction}
\end{equation}

\subsection{Pre-trained Encoder}
\label{sec:pretrain_encoder}
\begin{wrapfigure}{r}{0.5\textwidth}
\vskip -0.5cm
\centering
    \includegraphics[width=0.5\textwidth]{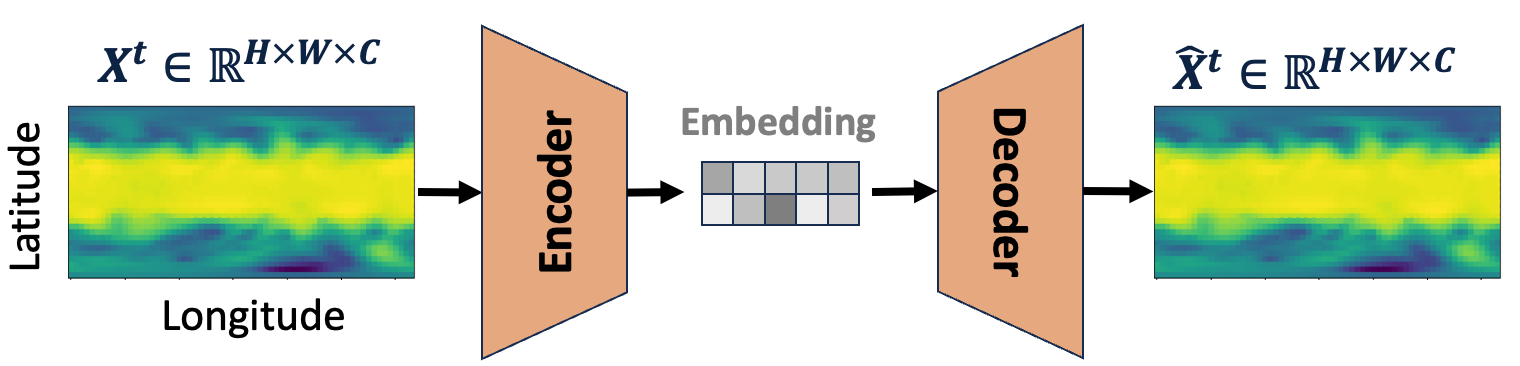}
\vspace{-4mm}
\caption{Autoencoder structure.}
\vskip -0.5cm
\label{fig:autoencoder}
\end{wrapfigure}
We learn an encoder by training an autoencoder network \citep{baldi2012autoencoders}.
An \texttt{Encoder} compresses the input at each time point into a latent-space representation, while \texttt{Decoder} reconstructs the input from the latent representation. 
After the encoder, $\mathcal{F}$, is trained, it can serve as a pre-trained representation learning model to project the original data into latent embedding in Eq. (\ref{eq:encoder}). Appendix \ref{sec:autoencoder_arch} provides more details.
\begin{equation}
    \tilde{Z}^{t-1:t}= \mathcal{F}(X^{t-1}, X^t)
    \label{eq:encoder}
\end{equation}

\subsection{Attention-based Denoiser Network}
\begin{wrapfigure}{l}{0.45\textwidth}
\vskip -0.5cm
\centering
    \includegraphics[width=0.45\textwidth]{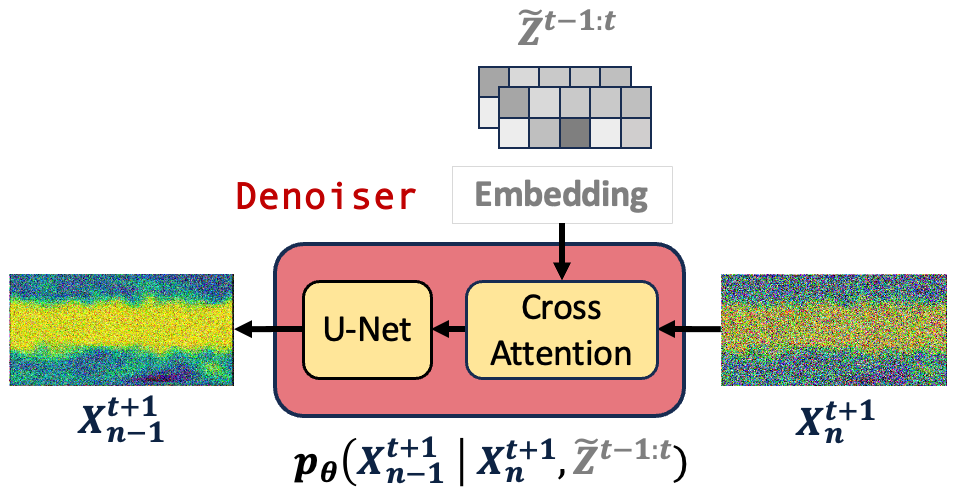}
\vspace{-5mm}
\caption{Attention-based denoiser structure.}
\vskip +0.3cm
\label{fig:denoiser}
\end{wrapfigure}
Our denoiser network consists of two blocks: cross-attention and U-net (as shown in Figure \ref{fig:denoiser}). 
Cross-attention mechanism \citep{hertz2022prompt} is employed to capture how past observations can contribute to the generation of future states.
The embedding of past observations, $\tilde{Z}^{t-1:t}$, and the noise data $X_n^{t+1}$ at diffusion step $n$, are projected to the same hidden dimension $d$ with the following transformation: 
\begin{small}
\begin{equation}
    Q=W_q \cdot X_n^{t+1}, K=W_k \cdot \tilde{Z}^{t-1:t}, V=W_v \cdot \tilde{Z}^{t-1:t},
\end{equation}
\end{small}
\noindent where $X_n^{t+1} \in \mathbb{R}^{(H \times W) \times C}$ and $\tilde{Z}^{t-1:t} \in \mathbb{R}^{(H \times W) \times d_z}$. $W_q \in \mathbb{R}^{d \times C}, W_k \in \mathbb{R}^{d \times d_z}, W_v \in \mathbb{R}^{d \times d_z}$ are learnable projection matrices \citep{vaswani2017attention}. Then we implement the cross-attention mechanism by $\texttt{Attention(Q, K, V)} = \texttt{softmax}(\frac{QK^T}{\sqrt{d}})V$. A visual depiction of the cross-attention mechanism is in Appendix \ref{sec:cross_attention_arch}.

U-Net \citep{ronneberger2015u} is utilized to recover the data by removing the noise added at each diffusion step.
The \emph{skip connection} technique in U-Net concatenates feature maps from the encoder to the corresponding decoder layers, allowing the network to retain fine-grained information that might be lost during downsampling.
The detailed U-Net architecture is presented in Appendix \ref{sec:unet_arch}.

\subsection{Training Process}
The training procedure is shown in Algorithm \ref{alg:training}. Firstly, we pre-train an \texttt{encoder} to learn the condition embedding of the past observations. Subsequently, we inject it into our conditional diffusion model and train \CoDiCast~with the devised loss function:
\begin{equation}
    \mathcal{L}_{cond}(\theta) = \mathbb{E}_{X_0, \epsilon, n} \left\| \epsilon - \epsilon_\theta \left(X_{n}^{t+1}, n, \texttt{cond} \right) \right\|^2,
    \label{eq:condition_loss}
\end{equation}
where $X_{n}^{t+1} = \sqrt{\bar{\alpha}_n} X_{0}^{t+1} + \sqrt{1-\bar{\alpha}_n} \epsilon$, $\texttt{cond}=\mathcal{F}(X^{t-1:t})$, and $\epsilon_\theta$ is the denoiser in Figure \ref{fig:denoiser}. 
\subsection{Inference Process}
Algorithm \ref{alg:inference} describes the inference process. 
We first extract the conditional embedding representations, $\tilde{Z}^{t-1:t}$, by the pre-trained encoder, and then randomly generate a noise vector $X_N \sim \mathcal{N}(\mathbf{0}, \mathbf{I})$ of size $H \times W \times C$. 
The sampled noise vector, $X_N$, is autoregressively denoised along the reversed chain to predict the target until $n$ equals 1 ($\zeta$ is set to zero when $n = 1$), we obtain weather prediction $\hat{X}_0$ at the time $t+1$. 
Later, multi-step prediction can be implemented autoregressively - the output from the previous time step is the input while predicting the next step, as shown in Eq. (\ref{eq:multi_prediction}).
\begin{minipage}[t]{0.45\textwidth} 
    \begin{algorithm}[H]
    \caption{Training}
    \label{alg:training}
    \begin{small}
    \begin{algorithmic}[1]
    \STATE \textbf{Input}: Number of diffusion steps $N$, pre-trained encoder $\mathcal{F}$ \\
    \STATE \textbf{Output}: Trained denoising function $\epsilon(\cdot)$  \\
    \REPEAT
        \STATE $X^{t+1}_0 \sim q(X^{t+1}_0)$
        \STATE $n \sim \texttt{Uniform}({1, 2, \dots , N})$
        \STATE $\epsilon \sim \mathcal{N}(\mathbf{0}, \mathbf{I})$
        \STATE Get the past observations $X^{t-1}, X^t$
        \STATE Get embedding $\tilde{Z}^{t-1:t}=\mathcal{F}(X^{t-1}, X^t)$
        \STATE Take gradient descent step on:
        $$\nabla_\theta \left\| \epsilon - \epsilon_\theta \left( X_{n}^{t+1}, n, \tilde{Z}^{t-1:t} \right) \right\|^2$$
    \UNTIL{converged}
    \end{algorithmic}
    \end{small}
    \end{algorithm}
\end{minipage}
\hfill
\begin{minipage}[t]{0.5\textwidth} 
    \begin{algorithm}[H]
    \caption{Inference}
    \label{alg:inference}
    \begin{small}
    \begin{algorithmic}[1]
    \STATE \textbf{Input}: Number of diffusion steps $N$, pre-trained encoder $\mathcal{F}$, trained denoising network $\epsilon(\cdot)$, past observations $X^{t-1}, X^t$ \\
    \STATE \textbf{Output}: Inference target $X_0^{t+1}$  \\
    \STATE Get embedding $\tilde{Z}^{t-1:t}=\mathcal{F}(X^{t-1}, X^t)$
    \STATE $X_N \sim \mathcal{N}(\mathbf{0}, \mathbf{I})$
    \FOR{$n = N, \dots, 1$}
        \STATE $\mathbf{\zeta} \sim \mathcal{N}(\mathbf{0}, \mathbf{I}) \text{ if } n \geq 1, \text{ else } \mathbf{\zeta} = 0$
        \STATE $X_{n-1}^{t+1} = \frac{1}{\sqrt{\alpha_n}} \left( X_n^{t+1} - \frac{1 - \alpha_n}{\sqrt{1 - \bar{\alpha}_n}} \epsilon_\theta (X_n^{t+1}, n, \tilde{Z}^{t-1:t} \right) + \sigma_n \mathbf{\zeta}$
    \ENDFOR
    \RETURN $X_0^{t+1}$
    \end{algorithmic}
    \end{small}
    \end{algorithm}
\end{minipage}

\subsection{Ensemble Forecast}
\label{sec:ensemble_forecast}
To enhance the reliability of weather forecasts, \emph{ensemble forecast} strategy is often employed to capture the variability among forecasts by separately running multiple deterministic models, e.g., ensemble forecast suite (ENS) \citep{buizza2008comparison}.
In our approach, since \CoDiCast~is a probabilistic model that can generate a distribution of future weather scenarios rather than a single prediction, following \citep{price2023gencast}, we run the trained \CoDiCast~multiple times to get the ensemble instead. 
More specifically, by integrating both initial conditions and noise sampled from a Gaussian distribution, \CoDiCast~implements the ensemble forecast through multiple stochastic samplings during inference, capturing a range of possible forecasts for the uncertainty quantification.
\section{Experiments}
\label{sec:exp}
\subsection{Dataset and Baselines}
\label{sec:data_intro}
\paragraph{Dataset.} 
ERA5 \citep{hersbach2020era5} is a publicly available atmospheric reanalysis dataset provided by the European Centre for Medium-Range Weather Forecasts (ECMWF). 
Following the existing work \citep{verma2024climode}, we use the preprocessed $5.625^\circ$ resolution ($32\times64$) and 6-hour increment ERA5 dataset from WeatherBench \citep{rasp2020weatherbench}. 
We downloaded 5 variables for the globe: geopotential at 500 hPa pressure level (\texttt{Z500}), atmospheric temperature at 850 hPa pressure level (\texttt{T850}), ground temperature (\texttt{T2m}), 10 meter U wind component (\texttt{U10}) and 10 meter V wind component (\texttt{V10}). 
More details can be found in Table \ref{tab:our_data} in Appendix \ref{sec:data_summary}. 
\paragraph{Baselines.}
We comprise the following baselines: ClimODE \citep{verma2024climode}: a spatiotemporal continuous-time model that incorporates the physic knowledge of atmospheric \emph{advection} over time; ClimaX \citep{nguyen2023climax}: a state-of-the-art vision Transformer-based method trained on the same dataset (without pre-training that is used in the original paper); FourCastNet \citep{pathak2022fourcastnet}: a global data-driven weather model using adaptive Fourier neural operators;
Neural ODE \citep{chen2018neural}: an ODE network that learns the time derivatives as neural networks by solving an ordinary differential equation;
Integrated Forecasting System IFS \citep{rasp2020weatherbench}: a global numerical weather prediction (NWP) system, integrating multiple advanced physics-based models to deal with more meteorological variables across multiple altitudes. 
Our study focuses solely on a subset of these variables due to the limited computational resources, with IFS serving as the gold standard.
For a fair comparison, all ML models use the same data set described in Section \ref{sec:data_intro}.

\subsection{Experiments Design}
We use data between 2006 and 2015 as the training set, data in 2016 as the validation set, and data between 2017 and 2018 as the testing set. 
We assess the global weather forecasting capabilities of our method \CoDiCast~by predicting the weather at a future time $t+\Delta t$ ($\Delta t$ = 6 to 144 hours) based on the past two time units. 
To quantify the uncertainty in weather prediction, we generate an ``ensemble'' forecast by running \CoDiCast~five times during the inference phase. 

\vspace{-1mm}
\paragraph{Training.} We first pretrain an \texttt{encoder} with the \texttt{Autoencoder} architecture. For the diffusion model, we used U-Net as the denoiser network with 1000 diffusion/denoising steps. The architecture is similar to that of DDPM \citep{ho2020denoising} work.
We employ four U-Net units for both the downsampling and upsampling processes.
Before training, we apply Max-Min normalization \citep{ali2014data} to scale the input data within the range $[0, 1]$, mitigating potential biases stemming from varying scales \citep{shi2023explainable}. 
\texttt{Adam} was used as the optimizer, where the learning rate $=2e^{-4}$, decay steps $=10000$, decay rate $=0.95$. 
The batch size and number of epochs were set to 64 and 800 respectively. 
More training details and model configurations are in Appendix \ref{sec:training_details}.

\vspace{-2mm}
\noindent \paragraph{Evaluation Metrics.} Following \citep{verma2024climode}, we use latitude-weighted Root Mean Square Error (RMSE) and Anomaly Correlation Coefficient (ACC) as deterministic metrics.
RMSE measures the average difference between values predicted by a model and the actual values. ACC is the correlation between prediction anomalies relative to climatology and ground truth anomalies relative to climatology. It is a critical metric in climate science to evaluate the model's performance in capturing unusual weather or climate events.
Appendix \ref{sec:metric} contains the formulas of these metrics.

\subsection{Quantitative Evaluation}
\label{sec:performance}
\paragraph{Accuracy.} We compare different models in forecasting five primary meteorological variables as described in Section \ref{sec:data_intro}. 
Table \ref{tab:global_score} shows that \CoDiCast~presents superior performance over other MLWP baselines in terms of RMSE and ACC metrics, demonstrating diffusion models can capture the weather dynamics and make predictions accurately. 
However, there is still room for \CoDiCast~to be improved compared with the gold-standard \texttt{IFS} model which integrates more meteorological variables. 
\begin{table}[ht]
\centering
\small
\caption{Latitude-weighted RMSE ($\downarrow$) and ACC ($\uparrow$) comparison on global weather forecasting. The results of NODE, ClimaX, and ClimODE models are from the ClimODE paper \citep{verma2024climode}. N/A represents the values that are not available from their paper.
ForeCastNet was re-trained with their code \citep{pathak2022fourcastnet}. We employ the Monte Carlo Dropout approach \citep{gal2016dropout} during the inference to compute the uncertainty. We mark the scores in \textbf{bold} if our model \CoDiCast~performs the best among MLWP methods. }
\vspace{+2mm}
\resizebox{0.999\columnwidth}{!}{
    \begin{tabular}{lc|lllll|l|ccccc|c}
    \toprule
        \multirow{2}{*}{Variable} & \multirow{2}{*}{\makecell[c]{Lead time\\(Hours)}}  & \multicolumn{6}{c|}{RMSE ($\downarrow$)} & \multicolumn{6}{c}{ACC ($\uparrow$)} \\
        \cmidrule(lr){3-8} \cmidrule(lr){9-14}
        & & NODE & ClimaX & ForeCastNet & ClimODE & CoDiCast & IFS               & NODE & ClimaX & ForeCastNet & ClimODE & CoDiCast & IFS \\
    \midrule
        \multirow{5}{*}{Z500} 
         & 6   & 300.6 & 247.5 & 222.7\textcolor{gray}{$\pm18.1$}    &102.9\textcolor{gray}{$\pm9.3$}   &\textbf{73.1}\textcolor{gray}{$\pm6.7$}   & 26.9   & 0.96 & 0.97 & 0.97   & 0.99 &\textbf{0.99}   & 1.00\\
         & 12  & 460.2 & 265.3 & 310.9\textcolor{gray}{$\pm22.7$}    &134.8\textcolor{gray}{$\pm12.3$}  &\textbf{114.2}\textcolor{gray}{$\pm8.9$}  & 33.8   & 0.88 & 0.96 & 0.95   & 0.99 &\textbf{0.99}   & 0.99\\
         & 24  & 877.8 & 364.9 & 402.6\textcolor{gray}{$\pm27.3$}    &193.4\textcolor{gray}{$\pm16.3$}  &\textbf{186.5}\textcolor{gray}{$\pm11.8$} & 51.0   & 0.70 & 0.93 & 0.92   & 0.98 &\textbf{0.98}   & 0.99\\ 
         & 72  & N/A   & 687.0 & 755.3\textcolor{gray}{$\pm45.8$}    &478.7\textcolor{gray}{$\pm48.5$}  &\textbf{451.6}\textcolor{gray}{$\pm39.5$} & 123.2  & N/A  & 0.73 & 0.75   & 0.88 &\textbf{0.92}   & 0.98\\
         & 144 & N/A   & 801.9 & 956.1\textcolor{gray}{$\pm59.1$}    &783.6\textcolor{gray}{$\pm37.3$}  &\textbf{757.5}\textcolor{gray}{$\pm42.8$} & 398.7  & N/A  & 0.58 & 0.64   & 0.61 &\textbf{0.78}   & 0.86\\
    \midrule
        \multirow{5}{*}{T850} 
         & 6   & 1.82 & 1.64 & 1.75\textcolor{gray}{$\pm0.16$}    &1.16\textcolor{gray}{$\pm0.06$}  &\textbf{1.02}\textcolor{gray}{$\pm0.05$}      & 0.69  & 0.94 & 0.94 & 0.94   & 0.97 &\textbf{0.99}    & 0.99\\
         & 12  & 2.32 & 1.77 & 2.15\textcolor{gray}{$\pm0.20$}    &1.32\textcolor{gray}{$\pm0.13$}  &\textbf{1.26}\textcolor{gray}{$\pm0.10$}      & 0.75  & 0.85 & 0.93 & 0.92   & 0.96 &\textbf{0.99}    & 0.99\\
         & 24  & 3.35 & 2.17 & 2.51\textcolor{gray}{$\pm0.27$}    &1.55\textcolor{gray}{$\pm0.18$}  &\textbf{1.52}\textcolor{gray}{$\pm0.16$}      & 0.87  & 0.72 & 0.90 & 0.89   & 0.95 &\textbf{0.97}    & 0.99\\
         & 72  & N/A  & 3.17 & 3.69\textcolor{gray}{$\pm0.34$}    &2.58\textcolor{gray}{$\pm0.16$}  &\textbf{2.54}\textcolor{gray}{$\pm0.14$}      & 1.15  & N/A  & 0.76 & 0.77   & 0.85 &\textbf{0.93}    & 0.96\\
         & 144 & N/A  & 3.97 & 4.29\textcolor{gray}{$\pm0.42$}    &3.62\textcolor{gray}{$\pm0.21$}  &\textbf{3.61}\textcolor{gray}{$\pm0.19$}      & 2.23  & N/A  & 0.69 & 0.71   & 0.77 &\textbf{0.85}    & 0.81\\
    \midrule
        \multirow{5}{*}{T2m} 
         & 6   & 2.72 & 2.02 & 2.05\textcolor{gray}{$\pm0.18$}    &1.21\textcolor{gray}{$\pm0.09$}  &\textbf{0.95}\textcolor{gray}{$\pm0.07$}      & 0.69  & 0.82 & 0.92 & 0.94   & 0.97 &\textbf{0.99}    & 0.99\\
         & 12  & 3.16 & 2.26 & 2.49\textcolor{gray}{$\pm0.21$}    &1.45\textcolor{gray}{$\pm0.10$}  &\textbf{1.21}\textcolor{gray}{$\pm0.07$}      & 0.77  & 0.68 & 0.90 & 0.92   & 0.96 &\textbf{0.99}    & 0.99\\
         & 24  & 3.86 & 2.37 & 2.78\textcolor{gray}{$\pm0.26$}    &1.40\textcolor{gray}{$\pm0.09$}  &1.45\textcolor{gray}{$\pm0.07$}               & 1.02  & 0.79 & 0.89 & 0.91   & 0.96 &\textbf{0.99}    & 0.99\\
         & 72  & N/A  & 2.87 & 3.77\textcolor{gray}{$\pm0.32$}    &2.75\textcolor{gray}{$\pm0.49$}  &\textbf{2.39}\textcolor{gray}{$\pm0.37$}      & 1.26  & N/A  & 0.83 & 0.85   & 0.85 &\textbf{0.96}    & 0.96\\
         & 144 & N/A  & 3.38 & 4.39\textcolor{gray}{$\pm0.41$}    &3.30\textcolor{gray}{$\pm0.23$}  &3.45\textcolor{gray}{$\pm0.22$}               & 1.78  & N/A  & 0.83 & 0.81   & 0.79 &\textbf{0.91}    & 0.82\\
    \midrule
        \multirow{5}{*}{U10} 
         & 6   & 2.30 & 1.58 & 1.98\textcolor{gray}{$\pm0.17$}    &1.41\textcolor{gray}{$\pm0.07$}  &\textbf{1.24}\textcolor{gray}{$\pm0.06$}      & 0.61  & 0.85 & 0.92 & 0.87   & 0.91 &\textbf{0.95}    & 0.98\\
         & 12  & 3.13 & 1.96 & 2.58\textcolor{gray}{$\pm0.21$}    &1.81\textcolor{gray}{$\pm0.09$}  &\textbf{1.50}\textcolor{gray}{$\pm0.08$}      & 0.76  & 0.70 & 0.88 & 0.78   & 0.89 &\textbf{0.93}    & 0.98\\
         & 24  & 4.10 & 2.49 & 3.02\textcolor{gray}{$\pm0.27$}    &2.01\textcolor{gray}{$\pm0.10$}  &\textbf{1.87}\textcolor{gray}{$\pm0.09$}      & 1.11  & 0.50 & 0.80 & 0.71   & 0.87 &\textbf{0.89}    & 0.97\\
         & 72  & N/A  & 3.70 & 4.17\textcolor{gray}{$\pm0.36$}    &3.19\textcolor{gray}{$\pm0.18$}  &\textbf{3.15}\textcolor{gray}{$\pm0.19$}      & 1.57  & N/A  & 0.45 & 0.41   & 0.66 &\textbf{0.71}    & 0.94\\
         & 144 & N/A  & 4.24 & 4.63\textcolor{gray}{$\pm0.45$}    &4.02\textcolor{gray}{$\pm0.12$}  &4.25\textcolor{gray}{$\pm0.15$}               & 3.04  & N/A  & 0.30 & 0.28   & 0.35 &\textbf{0.42}    & 0.72\\
    \midrule
        \multirow{5}{*}{V10} 
         & 6   & 2.58 & 1.60 & 2.16\textcolor{gray}{$\pm0.19$}    &1.53\textcolor{gray}{$\pm0.08$}  &\textbf{1.30}\textcolor{gray}{$\pm0.06$}      & 0.61  & 0.81 & 0.92 & 0.86   & 0.92 &\textbf{0.95}    & 1.00\\
         & 12  & 3.19 & 1.97 & 2.73\textcolor{gray}{$\pm0.23$}    &1.81\textcolor{gray}{$\pm0.12$}  &\textbf{1.56}\textcolor{gray}{$\pm0.09$}      & 0.79  & 0.61 & 0.88 & 0.76   & 0.89 &\textbf{0.93}    & 0.99\\
         & 24  & 4.07 & 2.48 & 3.15\textcolor{gray}{$\pm0.28$}    &2.04\textcolor{gray}{$\pm0.10$}  &\textbf{1.94}\textcolor{gray}{$\pm0.14$}      & 1.33  & 0.35 & 0.80 & 0.68   & 0.86 &\textbf{0.89}    & 1.00\\
         & 72  & N/A  & 3.80 & 4.26\textcolor{gray}{$\pm0.34$}    &3.30\textcolor{gray}{$\pm0.22$}  &\textbf{3.18}\textcolor{gray}{$\pm0.19$}      & 1.67  & N/A  & 0.39 & 0.38   & 0.63 &\textbf{0.68}    & 0.93\\
         & 144 & N/A  & 4.42 & 4.64\textcolor{gray}{$\pm0.45$}    &4.24\textcolor{gray}{$\pm0.10$}  &\textbf{4.21}\textcolor{gray}{$\pm0.18$}      & 3.26  & N/A  & 0.25 & 0.27   & 0.32 &\textbf{0.37}    & 0.71\\
    \bottomrule
    \end{tabular}
}
\label{tab:global_score}
\end{table}
%
%
%
%
\paragraph{Uncertainty.}
\begin{wrapfigure}{r}{0.45\textwidth}
\vskip -0.4cm
\centering
    \includegraphics[width=0.45\textwidth]{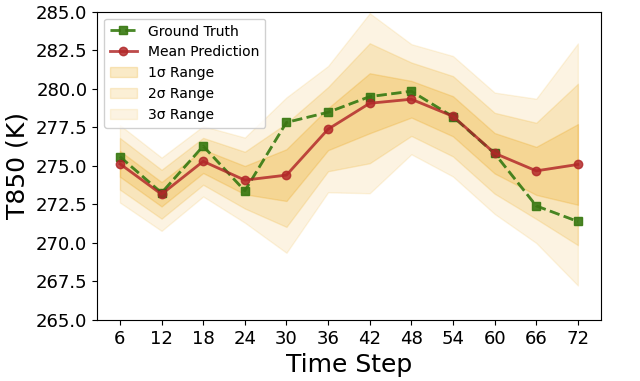}
\vspace{-6mm}
\caption{Forecast with confidence intervals.}
\vskip -0.3cm
\label{fig:forecast_uncertainty}
\end{wrapfigure}
The gray-shaded error range in Table \ref{tab:global_score} represents model uncertainty, with fluctuations remaining within 10\% of the ground truth scale. This indicates the robustness of ML models. However, \texttt{ForeCastNet} predictions fluctuate relatively larger due to the sensitive selection of the dropout rate.
We provide a case study of \CoDiCast~forecast for 72 hours with uncertainty quantification in Figure \ref{fig:forecast_uncertainty}.
It shows the mean prediction tracks the general trend of the ground truth and the uncertainty grows as the lead time increases.
The increasing uncertainty over time aligns with the common intuition that longer forecasting horizons make time series predictions more challenging.
Besides, we can tell that most actual values fall within the 1 or 2 standard deviations ($\sigma$) ranges, indicating these predictions are reasonably accurate.

%
\paragraph{Inference efficiency.} Generally, numerical weather prediction models (e.g., \texttt{IFS}) require around 50 minutes for the medium-range global forecast, while deterministic ML weather prediction models take less than 1 minute \citep{rasp2020weatherbench} but cannot model the weather uncertainty. 
\CoDiCast~needs about 3.6 minutes (see the last row in Table \ref{tab:rmse_vs_diffusion_step}) for the global weather forecast, potentially balancing the efficiency and accuracy with essential uncertainty quantification. 
The efficiency also depends on the model complexity.

\subsection{Qualitative Evaluation}
\begin{figure*}[ht]
\centering
    \includegraphics[width=0.99\textwidth]{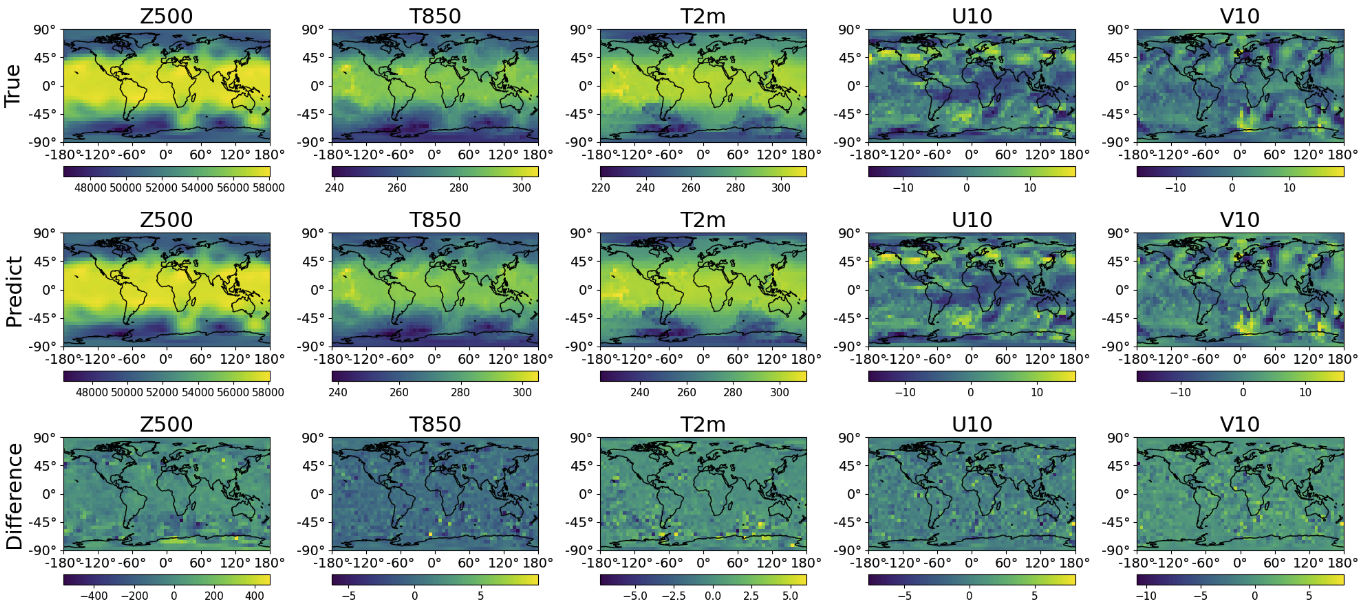}
\caption{Visualization of true and predicted values at 6 hours lead time.}
\label{fig:true_vs_pred_all_t6}
\end{figure*}

In Figure \ref{fig:true_vs_pred_all_t6}, we qualitatively evaluate the performance of \CoDiCast~on global forecasting tasks for all target variables, \texttt{Z500}, \texttt{T850}, \texttt{T2m}, \texttt{U10} and \texttt{V10} at the lead time of $6$ hours.
The first row is the ground truth of the target variable, the second row is the prediction and the last row is the difference between the model prediction and the ground truth.
From the scale of their color bars, we can tell that the error percentage is less than 3\% for variables \texttt{Z500}, \texttt{T850}, and \texttt{T2m}. Nevertheless, error percentages over $50\%$ exist for \texttt{U10} and \texttt{V10} even though only a few of them exist.
Furthermore, we observe that most higher errors appear in the high-latitude ocean areas, probably due to the sparse data nearby.
We provide visualizations for longer lead times (up to 3 days) in Appendix \ref{sec:vis_long}.
\subsection{Ablation Study}
\begin{figure*}[ht]
\centering
    \includegraphics[width=0.99\textwidth]{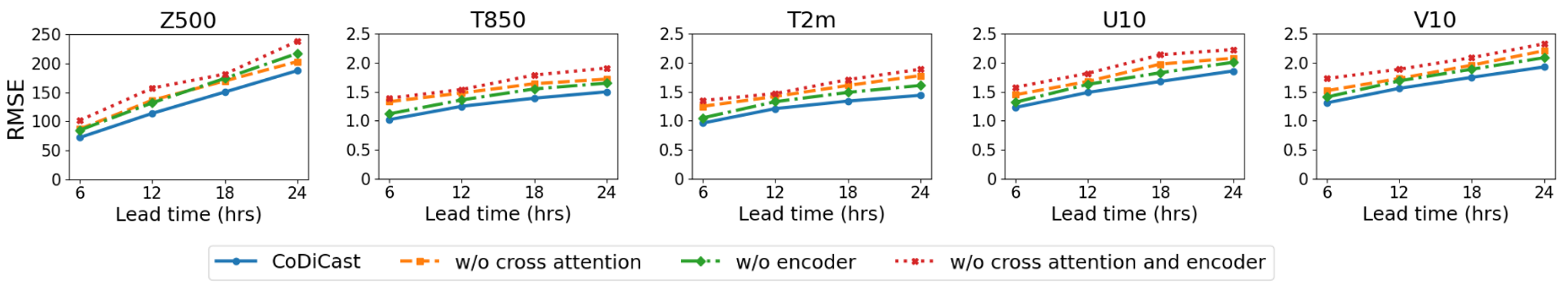}
\caption{Ablation study to study the effect of pre-trained \texttt{encoder} and \texttt{cross-attention}.}
\label{fig:ablation}
\end{figure*}

\CoDiCast~includes two important components: \emph{pre-trained encoder} and \emph{cross attention}. 
To study their effectiveness, we conduct an ablation study: (a) \textbf{No-encoder} directly considers past observations as conditions to diffusion model; (b) \textbf{No-cross-attention} simply concatenate the embedding and the noisy sample at each denoising step; (c) \textbf{No-encoder-cross-attention} concatenate the past observations and the noisy sample at each denoising step. Figure \ref{fig:ablation} shows the full \CoDiCast~consistently outperforms all other variants, verifying positive contributions of both components.

\subsection{Parameter Study}
\paragraph{Diffusion step.} We try various diffusion steps $N=\{250, 500, 750, 1000, 1500, 2000\}$. 
Table \ref{tab:rmse_vs_diffusion_step} shows that the accuracy improves as the number of diffusion steps increases when $N < 1000$, indicating that more intermediate steps are more effective in learning the imperceptible attributes during the denoising process.
However, when $1000<N<2000$, the accuracy remains approximately flat but the inference time keeps increasing linearly.
Considering the trade-off between accuracy and efficiency, we finally set $N=1000$ for all experiments.
\begin{figure*}[ht]
    \centering
    \includegraphics[width=0.99\textwidth]{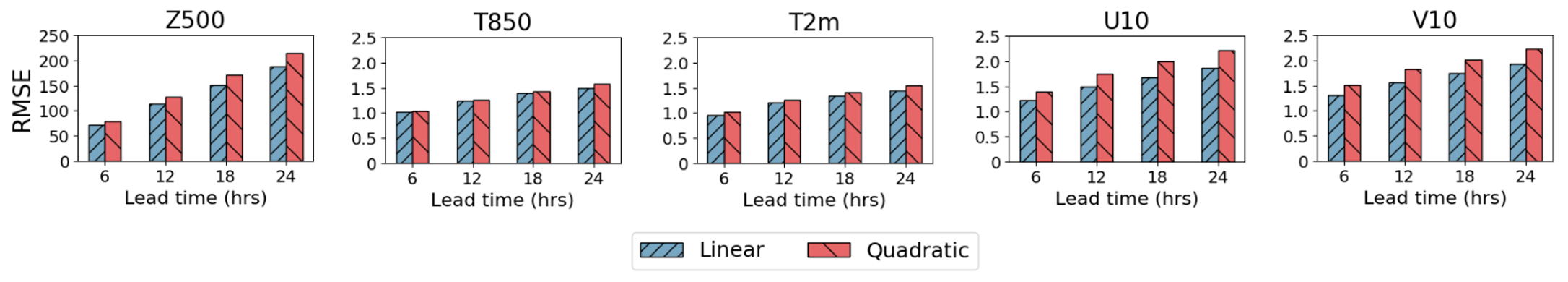}
    \caption{Effect of linear and quadratic variance scheduling methods.}
    \label{fig:variance_schedule}
\end{figure*}
\begin{wraptable}{r}{0.55\textwidth}
\vskip -0.5cm
\small
\caption{Latitude-weighted RMSE with various diffusion steps. We mark the lowest scores in bold font. The last row represents the inference time of \CoDiCast. }
\label{tab:rmse_vs_diffusion_step}
\vspace{+2mm}
\resizebox{.55\columnwidth}{!}{
    \begin{tabular}{lc|ccccccc}
    \toprule
        \multirow{2}{*}{Variable} & \multirow{2}{*}{\makecell[c]{Lead \\time}} & \multicolumn{6}{c}{Diffusion Step} \\
        \cmidrule(lr){3-8}
        & & 250 & 500 & 750 & 1000 & 1500 & 2000\\
    \midrule
        \multirow{1}{*}{Z500} 
         & 24 & 696.1 & 324.8 & 190.6   &\textbf{186.5} &193.5  &191.9 \\
    \midrule
        \multirow{1}{*}{T850} 
         & 24 & 3.88 & 2.38 & 1.53   &\textbf{1.52}     &1.56   &1.58 \\
    \midrule
        \multirow{1}{*}{T2m} 
         & 24 & 5.26 & 2.79 & 1.63   &\textbf{1.44}     &1.50   &1.53 \\
    \midrule
        \multirow{1}{*}{U10} 
         & 24 & 2.74 & 2.05 & \textbf{1.81}  &1.87      &1.99   &2.01 \\
    \midrule
        \multirow{1}{*}{V10} 
         & 24 & 2.43 & 2.11 & \textbf{1.89}  &1.94      &2.04   &2.06 \\
    \midrule
        \multicolumn{2}{l|}{Inference time (min)} 
              &$\sim1.1$ &$\sim1.9$  &$\sim2.8$  &$\sim3.6$  &$\sim6.5$   &$\sim8.3$  \\
    \bottomrule
    \end{tabular}
}
\vskip -0.5cm
\end{wraptable}
\vspace{-1.0cm}
\paragraph{Method for variance scheduling.} We use the same start and end variance value, $\beta$, as DDPM \citep{ho2020denoising} where $\beta \in [0.0001, 0.02]$. 
We study the effect of ``linear'' and ``quadratic'' variance scheduling in this section. 
The results are provided in Figure \ref{fig:variance_schedule}. It shows that the ``linear'' variance scheduling provides better performance than ``quadratic'' one for variables \texttt{Z500}, \texttt{T2m}, \texttt{U10}, and \texttt{V10}, while the performance of both ``linear'' and ``quadratic'' modes is roughly same for variable \texttt{T850}.

\section{Related Work}
\label{sec:relate}

\textbf{Physics-based Numerical Weather Prediction.} 
Numerical Weather Prediction (NWP) methods achieve weather forecasts by modeling the system of the atmosphere, land, and ocean with complex differential equations \citep{bauer2015quiet}. 
For example, High-Resolution Forecasts System (HRES) \citep{HRES2023ECMWF} forecasts possible weather evolution out to 10 days ahead. 
However, it is a deterministic NWP method that only provides a single forecast. 
To overcome the limitation of deterministic methods, the ensemble forecast suite (ENS) \citep{buizza2008comparison} was developed as an ensemble of 51 forecasts by the European Centre for Medium-Range Weather Forecasts (ECMWF). ENS provides a range of possible future weather states, allowing for investigation of the detail and uncertainty in the forecast.
Even if NWP ensemble forecasts effectively model the weather evolution, they exhibit sensitivity to structural discrepancies across models and high computational demands \citep{balaji2022general}.

\paragraph{ML-Based Weather Prediction.} 
Pangu \citep{bi2023accurate} employed three-dimensional transformer networks and Earth-specific priors to deal with complex patterns in weather data.
GraphCast \citep{lam2023learning} achieved medium-range weather prediction by utilizing an ``encode-process-decode'' configuration with each part implemented by graph neural networks (GNNs). GNNs perform effectively in capturing the complex relationship between a set of surface and atmospheric variables. A similar GNN-based work is \citep{keisler2022forecasting}. 
Fuxi \citep{chen2023fuxi} and Fengwu \citep{chen2023fengwu} also employ the ``encode-decode'' strategy but with the transformer-based backbone.
FourCastNet \citep{pathak2022fourcastnet} applied Vision Transformer (ViT) and Adaptive Fourier Neural Operators (AFNO), while ClimaX \citep{nguyen2023climax} also uses a ViT backbone but the trained model can be fine-tuned to various downstream tasks.
However, these models fall short in modeling the uncertainty of weather evolution \citep{jaseena2022deterministic} even though perturbations are added to initial conditions \citep{bulte2024uncertainty} and dropout methods \citep{gal2016dropout} are used to mimic the uncertainty.
Additionally, ClimODE \citep{verma2024climode} incorporated the physical knowledge and developed a continuous-time neural advection PDE weather model.

\paragraph{Diffusion Models.} 
Diffusion models \citep{ho2020denoising} have shown their strong capability in computer vision tasks, including image generation \citep{li2022srdiff}, image editing \citep{nichol2021glide}, semantic segmentation \citep{brempong2022denoising} and point cloud completion \citep{luo2021diffusion}.
Conditional diffusion models \citep{ho2022classifier} were later proposed to make the generation step conditioned on the current context or situation.
However, not many efforts have adopted diffusion models in global medium-range weather forecasting.
More recent research has focused on precipitation nowcasting \citep{asperti2023precipitation, gao2024prediff}, and are localized in their predictions. 
GenCast \citep{price2023gencast} is a recently proposed conditional diffusion-based ensemble forecasting for medium-range weather prediction.
However, their conditioning is to directly use the observations from the recent past, which is shown to be insufficient (see results in ablation study).
\section{Conclusions}
\label{sec:conc}
\vspace{-1mm}
In this work, we start with analyzing the limitations of current deterministic numerical weather prediction (NWP) and machine-learning weather prediction (MLWP) approaches---they either cause substantial computational cost or lack uncertainty quantification in the forecasts. 
To address these limitations, we propose a conditional diffusion model, \CoDiCast, which contains a conditional \emph{pre-trained encoder} and a \emph{cross-attention} component.
Quantitative and qualitative experimental results demonstrate it can simultaneously complete more accurate predictions than existing MLWP-based models and a faster inference than NWP-based models while being capable of providing uncertainty quantification compared to deterministic methods.
In conclusion, our model offers three key characteristics at the same time.


\noindent \textbf{Limitation and Future work.} We use low-resolution ($5.625^\circ$) data currently due to the relatively slow inference process of diffusion models compared to deterministic ML models. In the future, we will focus on accelerating diffusion models \citep{song2020denoising} to adapt the higher-resolution data.
Besides the meteorological numerical data, weather events are often recorded or reported in the form of text. We will study how to leverage LLMs to extract their implicit interactions \citep{li2024cllmate} and inject them into diffusion models to guide the generation process.

\section*{Acknowledgments}
This work is supported by the Institute for Geospatial Understanding through an Integrative Discovery Environment (I-GUIDE), which is funded by the National Science Foundation (NSF) under award number: 2118329. Any opinions, findings, conclusions, or recommendations expressed herein are those of the authors and do not necessarily represent the views of NSF.

\bibliography{iclr2025_conference}
\bibliographystyle{iclr2025_conference}

\appendix
\onecolumn
\section*{Appendix}

\section{Dataset}
\label{sec:data_summary}
We introduce a detailed description of the ERA5 dataset. As the predominant data source for learning and benchmarking weather prediction systems, the ERA5 reanalysis archive from the European Center for Medium-Range Weather Forecasting (ECMWF) provides reanalyzed data from 1979 onwards. 
This data is available on a $0.25^\circ \times 0.25^\circ$ global latitude-longitude grid of the Earth’s sphere, at hourly intervals, with different atmospheric variables at 37 different altitude levels and some variables on the Earth’s surface. The grid overall contains $721 \times 1440$ grid points for latitude and longitude, respectively.
Due to the limited computational resources, we used the preprocessed version of ERA5 from WeatherBench \citep{rasp2020weatherbench} in our work. 
This dataset\footnotemark[1] contains re-gridded ERA5 reanalysis data in three lower resolutions: $5.625^\circ$, $2.8125^\circ$, and $1.40625^\circ$. 
To guarantee fair comparison with the benchmarks \citep{verma2024climode}, we follow the \texttt{ClimODE} work and choose the $5.625^\circ$ resolution dataset for variables: geopotential at 500 hPa pressure level (\texttt{Z500}), atmospheric temperature at 850 hPa pressure level (\texttt{T850}), ground temperature (\texttt{T2m}), 10 meter U wind component (\texttt{U10}) and 10 meter V wind component (\texttt{V10}). 
\emph{Single} represents surface-level variables, and \emph{Atmospheric} represents time-varying atmospheric properties at chosen altitudes.
A sample at a certain time point can be represented by $X^t \in \mathbb{R}^{H \times W \times C}$ where $H \times W$ refers to the spatial resolution of data which depends on how densely we grid the globe over latitude and longitude, $C$ refers to the number of channels (i.e, weather variables). In our work, $H, W$, and $C$ are $32, 64$, and $5$, accordingly.
Notably, both \texttt{Z500} and \texttt{T850} are two popular verification variables for global weather prediction models, while \texttt{T2m}, \texttt{U10}, and \texttt{V10} directly pertain to human activities.
\footnotetext[1]{\url{https://github.com/pangeo-data/WeatherBench}}
\begin{table}[ht]
\small
\centering
\caption{Variable Information.}
\vspace{+2mm}
\label{tab:our_data}
\begin{tabular}{llcccccc}
    \toprule
    Type        & Variable              & Abbrev.   & ECMWF ID  & Levels    & Range                 & Unit  \\
    \midrule
    Single      & 2 metre temperature   & T2m       & 167       &           & $[193.1, 323.6]$      & $K$   \\
    Single      & 10 metre U wind       & U10       & 165       &           & $[-37.3, 30.2]$       & $m/s$ \\
    Single      & 10 metre V wind       & V10       & 166       &           & $[-31.5, 32.5]$       & $m/s$ \\
    Atmospheric & Geopotential          & Z         & 129       & 500       & $[43403.6, 59196.9]$  & $m^2/s^2$ \\
    Atmospheric & Temperature           & T         & 130       & 850       & $[217.9, 313.3]$      & $K$   \\
    \bottomrule
\end{tabular}
\end{table}

\section{Model Architecture}
\label{sec:model_details}
We present the detailed architectures of the autoencoder, cross-attention block, and U-Net model used in our work. 
Meanwhile, we also illustrate how we organize the input data and how they flow through different machine-learning model blocks.
We recommend readers check out Figures \ref{fig:framework}, \ref{fig:autoencoder}, and \ref{fig:denoiser} while looking into the following architectures.

\subsection{Autoencoder}
\label{sec:autoencoder_arch}
We train an autoencoder model consisting of two main parts: an encoder and a decoder.
The encoder compresses the input to feature representation (embedding) in the latent space. 
The decoder reconstructs the input from the latent space. After training, the pre-trained encoder can be extracted to generate embedding for input data.
In our work, the convolutional autoencoder architecture is designed for processing spatiotemporal weather data at a time point, $t$, represented as $X^t \in \mathbb{R}^{H \times W \times C}$. 
The encoder consists of a series of convolutional layers with $2\times2$ filters, each followed by a \texttt{ReLU} activation function. 
The layers have $32$, $128$, $256$, and $512$ filters, respectively, allowing for a progressive increase in feature depth, thereby capturing essential patterns in the data. 
The decoder starts with $512$ filters and reduces the feature depth through layers with $256$ and $128$ filters, each followed by \texttt{ReLU} activations. 
This design ensures the reconstruction of the input data while preserving the learned features, enabling the model to extract meaningful embeddings that encapsulate the spatiotemporal characteristics of the input.
\begin{figure*}[ht]
\centering
    \includegraphics[width=0.8\textwidth]{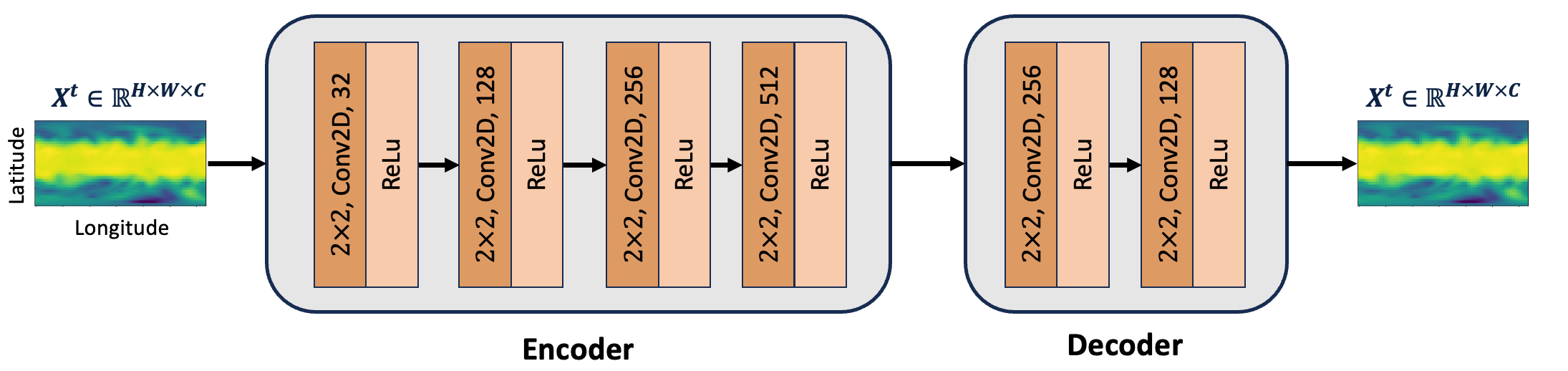}
\caption{Architecture of the Autoencoder model.}
\label{fig:autoencoder_arch}
\end{figure*}

\subsection{Cross-Attention}
\label{sec:cross_attention_arch}
The cross-attention is used to learn the interaction between past observations and the noisy data at each diffusion step. We consider the past observations as the conditions to guide the diffusion models during generation.
Given the weather states in the past two time points, $X^{T \times H \times W \times C}$, we utilize the pre-trained encoder to learn the embedding from each time point, $X^{T \times H \times W \times d_e}$. 
To better use the \emph{attention} mechanism, we first reshape it to $X^{(H * W) \times (T*d_e)}$ and convert it to key and value matrices: $K \in \mathbb{R}^{(H * W) \times d_k}$ and $V \in \mathbb{R}^{(H * W) \times d_v}$.
We consider the noisy sample at each diffusion step, $X_n \in \mathbb{R}^{T \times H \times W \times C}$, as a query. 
It is transformed to $Q\in \mathbb{R}^{(H * W) \times d_q}$. Then, the \emph{cross-attention} mechanism is implemented by $Attention(Q, K, V) = softmax(\frac{QK^T}{\sqrt{d}}) \cdot V$.
In our work, we set $d_q = d_k = d_v = d = 64$ where $d$ is the projection embedding length.
\begin{figure*}[ht]
\centering
    \includegraphics[width=0.6\textwidth]{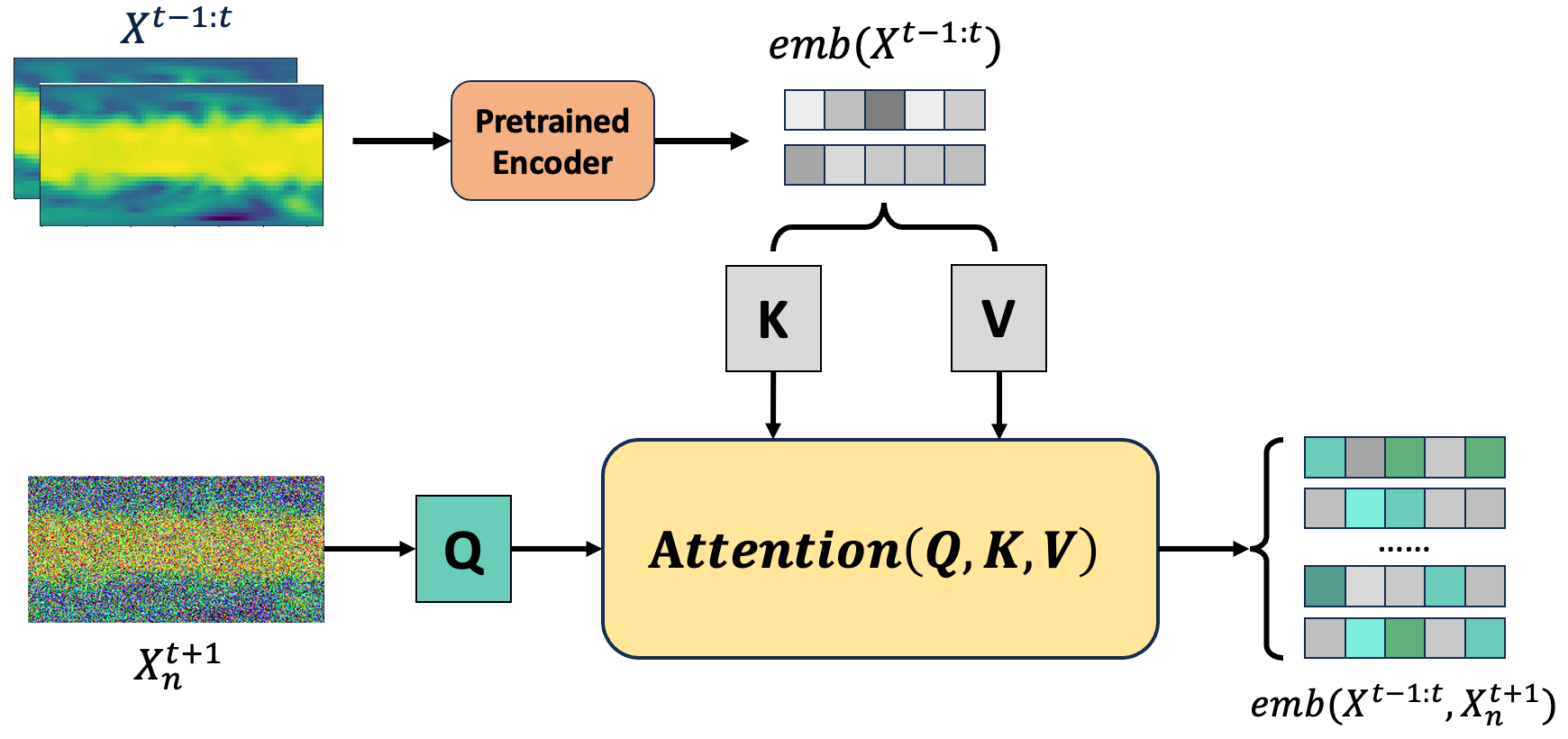}
\caption{Architecture of the cross-attention block.}
\label{fig:cross_attention}
\end{figure*}

\subsection{U-Net}
\label{sec:unet_arch}
Our U-Net architecture is similar to DDPM \citep{ho2020denoising} but with necessary changes to adapt to the problems in this work. 
Each U-Net unit comprises two ResNet blocks \citep{he2016deep} and a convolutional up/downsampling block. 
Self-attention was included between the convolution blocks once we reached a specific resolution ($4\times8, 2\times4$), represented in blue arrows. We employ four U-Net units for both the downsampling and upsampling processes. 
We use \emph{MaxPooling} in the downsampling units where the channel dimension is $64 \times j$ ($j=\{1,2,3,4\}$ refers to the layer index). 
The upsampling units follow the reverse order. We set the upsampling factor as 2 and the ``nearest" interpolation.
We used the \texttt{swish} activation function throughout the network. 
We also had \texttt{GroupNormalization} layer for more stable training where the number of groups for Group Normalization is 8. Group Normalization divides the channels into groups and computes within each group the mean and variance for normalization. 

Notably, for the target variable, $X^{t+1}$ at the $n$ diffusion step, the input to U-Net involves the mixture embedding of past weather states, $X^{t-1:t}$, and the noisy sample from the last diffusion step, $X^{t+1}_{n}$. The mixture embedding is obtained by the cross-attention mechanism described above.
The channel dimension output is five because of five weather variables of interest to predict. This is achieved by a convolutional layer with a $1 \times 1$ kernel.
\begin{figure*}[ht]
\centering
    \includegraphics[width=0.89\textwidth]{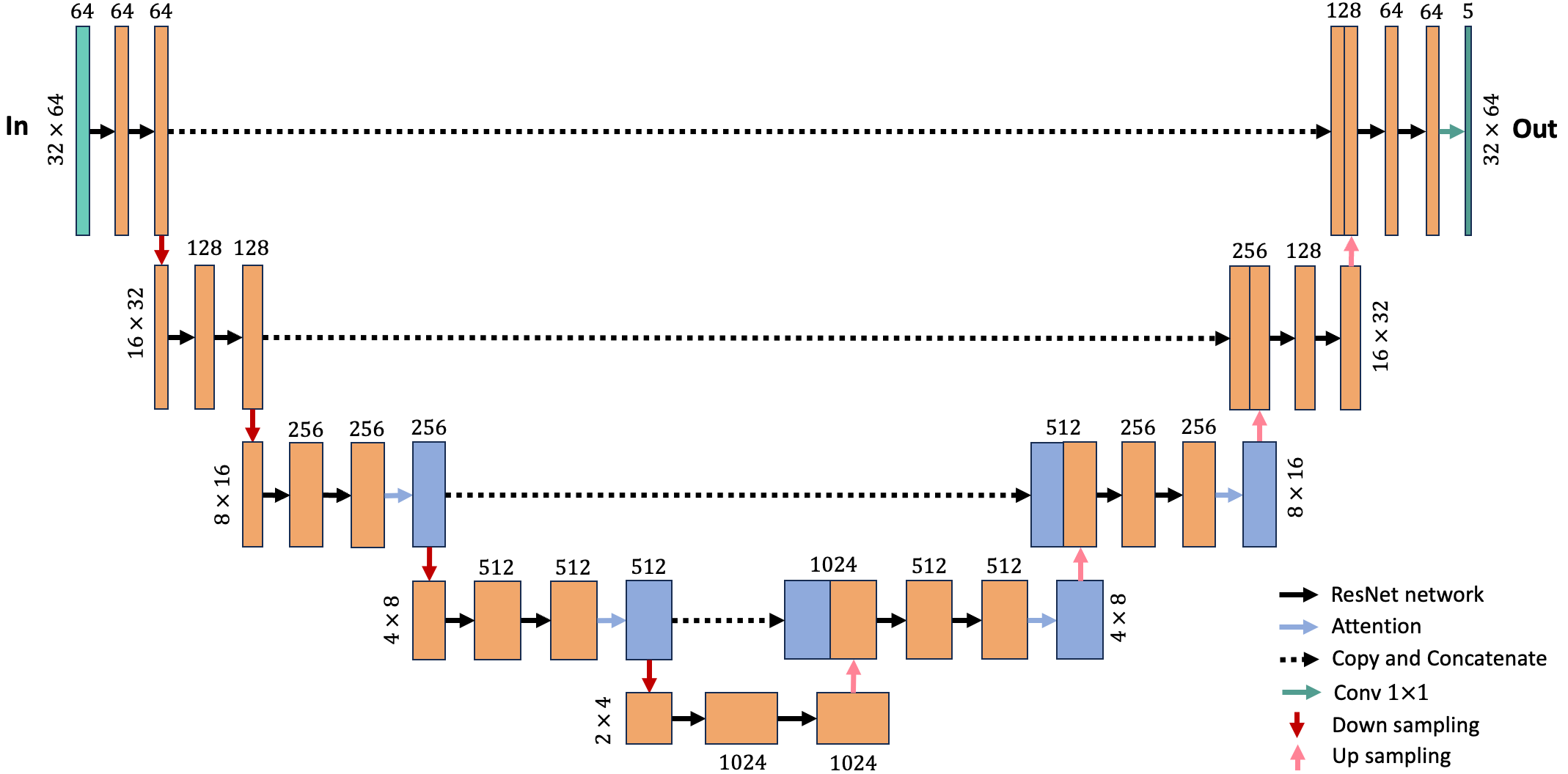}
\caption{Architecture of the U-Net model.}
\label{fig:unet_arch}
\end{figure*}

\section{Training Details}
\label{sec:training_details}
We provide the hyperparameters for training our model \CoDiCast, which includes pre-training \texttt{Autoencoder} and training the \texttt{denoiser} network. 
Since it is more helpful to find the minimum loss if using a decayed learning rate as the training progresses, we applied an exponential decay function to an optimizer step given a provided initial learning rate.
\begin{table}[ht]
\small
\centering
\caption{Hyperparameters of Training Autoencoder.}
\vspace{+2mm}
\label{tab:train_autoencoder}
\small
    \begin{tabular}{lcc}
        \toprule
        Abbreviation   & Training Autoencoder & Training Denoiser   \\
        \midrule
        Epochs         & 100                            & 800  \\
        Batch\_size    & 128                            & 256   \\
        Learning\_rate & 1e-4                           & 2e-4  \\
        Decay\_steps   & 10000                          & 10000  \\
        Decay\_rate    & 0.95                           & 0.95  \\
        \bottomrule
    \end{tabular}
\end{table}

\section{Evaluation Metrics}
\label{sec:metric}
\paragraph{Root Mean Square Error.} Following \citep{verma2024climode}, we assess the model performance using latitude-weighted Root Mean Square Error (RMSE).
RMSE measures the average difference between values predicted by a model and the actual values. 
\begin{equation*}
\text{RMSE} = \frac{1}{M} \sum_{m=1}^{M} \sqrt{\frac{1}{H \times W} \sum_{h=1}^{H} \sum_{w=1}^{W} L(h) (\tilde{X}_{m,h,w} - X_{m,h,w})^2},
\end{equation*}
where $L(h)=\frac{1}{H}cos(h)\sum_{h'}^{H} cos(h')$ is the latitude weight and $M$ represents the number of test samples.
\paragraph{Anomaly Correlation Coefficient.} ACC is the correlation between prediction anomalies $\tilde{X}'$ relative to climatology and ground truth anomalies $\hat{X}$ relative to climatology.
ACC is a critical metric in climate science to evaluate the model's performance in capturing unusual weather or climate events.
\begin{equation*}
\text{ACC} = \frac{\sum_{m,h,w} L(h) \tilde{X}'_{m,h,w} {X'}_{m,h,w}}{\sqrt{\sum_{m,h,w} L(h) \tilde{X}_{m,h,w}^{'2} \cdot \sum_{m,h,w} L(h) {X}_{m,h,w}^{'2}}},
\end{equation*}
where observed and forecasted anomalies $X' = X - C, \tilde{X}' = \tilde{X} - C$, and climatology $C=\frac{1}{M} \sum_m X$ is the temporal mean of the ground truth data over the entire test set.

\section{Visualizing Prediction with longer lead times}
\label{sec:vis_long}
We provide the forecast at longer lead times (i.e., $24, 48, 72, 144$ hours). The first row is the ground truth of the target variable, the second row is the prediction of \CoDiCast~and the last row is the difference between the model prediction and the ground truth. 
\subsection{Short range weather forecasting}
\label{sec:short_cast}
Short-range weather forecasting at the 24-hour lead time for all target variables.
\begin{figure}[ht!]
    \centering
    \includegraphics[width=0.99\textwidth]{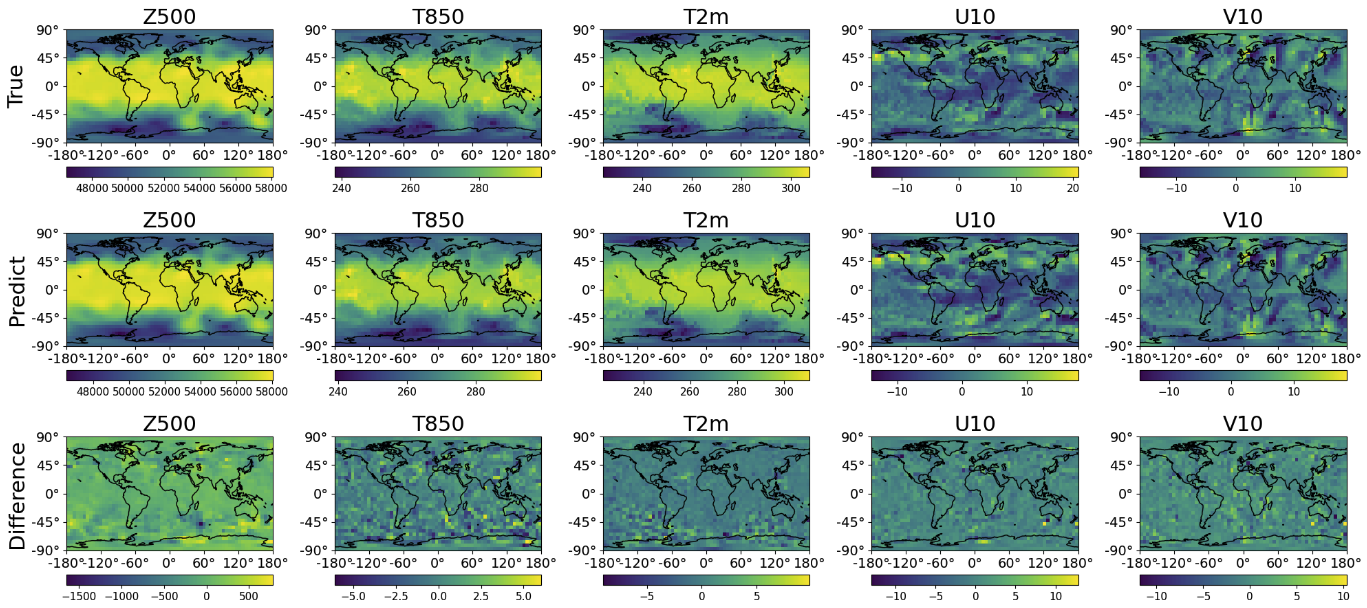}
    \caption{Visualizations of true and predicted values of all five variables at 24 hours lead time.}
    \label{fig:true_vs_pred_all_t24}
\end{figure}

\subsection{Medium-range weather forecasting}
\label{sec:medium_cast}
Medium-range weather forecasting at the 48-hour lead time for all target variables.
\begin{figure}[ht!]
    \centering
    \includegraphics[width=0.99\textwidth]{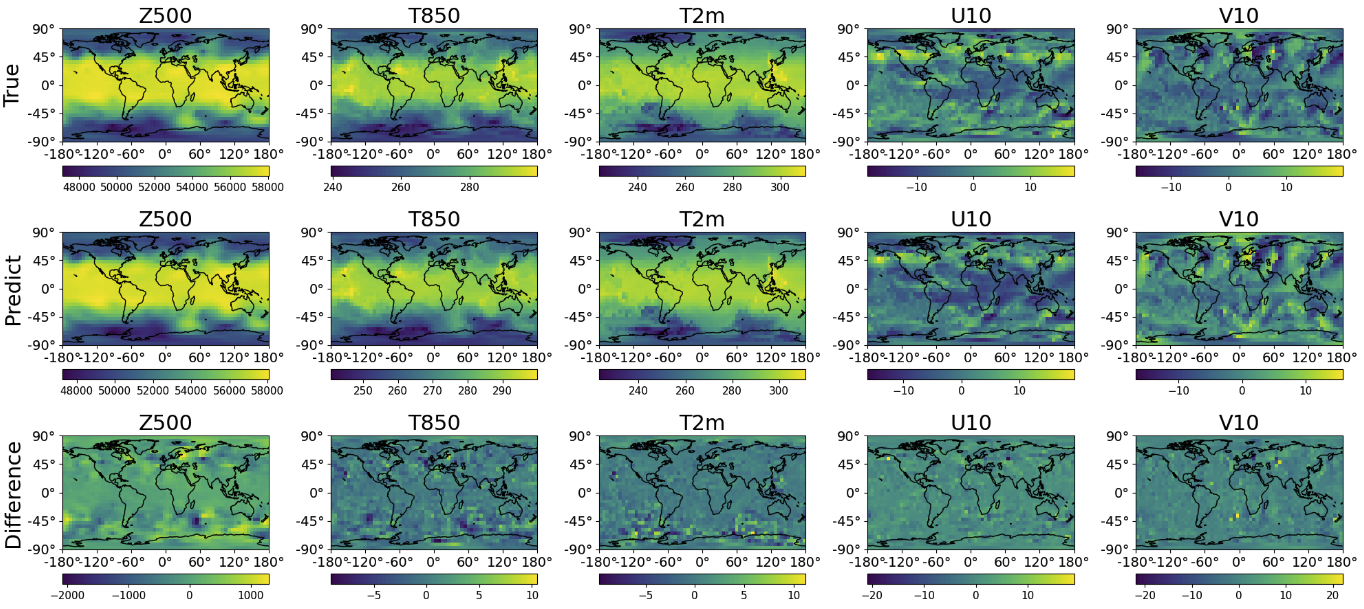}
    \caption{Visualizations of true and predicted values of all five variables at 48 hours lead time.}
    \label{fig:true_vs_pred_all_t48}
\end{figure}

\newpage
\subsection{Medium-range weather forecasting}
\label{sec:medium_cast1}
Longer-range weather forecasting at the 72-hour lead time for all target variables.
\begin{figure}[ht!]
    \centering
    \includegraphics[width=0.99\textwidth]{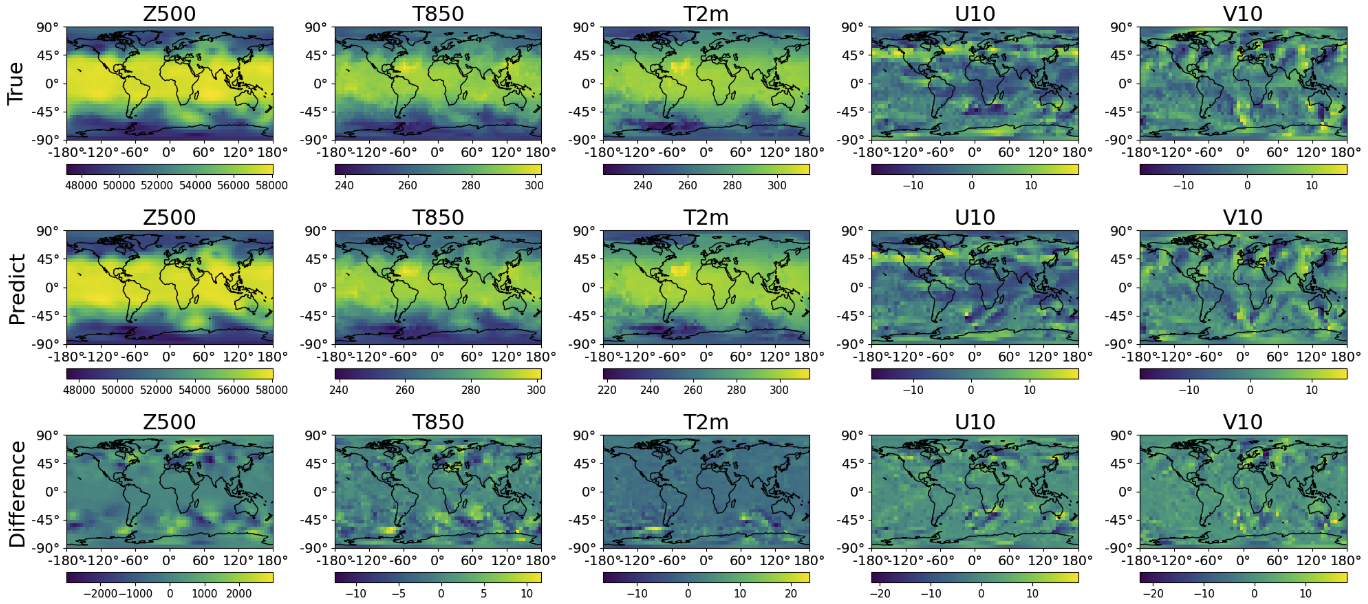}
    \caption{Visualizations of true and predicted values of all five variables at 72 hours lead time.}
    \label{fig:true_vs_pred_all_t72}
\end{figure}

\subsection{Long-range weather forecasting}
\label{sec:long_cast}
Longer-range weather forecasting at the 144-hour lead time for all target variables.
\begin{figure}[ht!]
    \centering
    \includegraphics[width=0.99\textwidth]{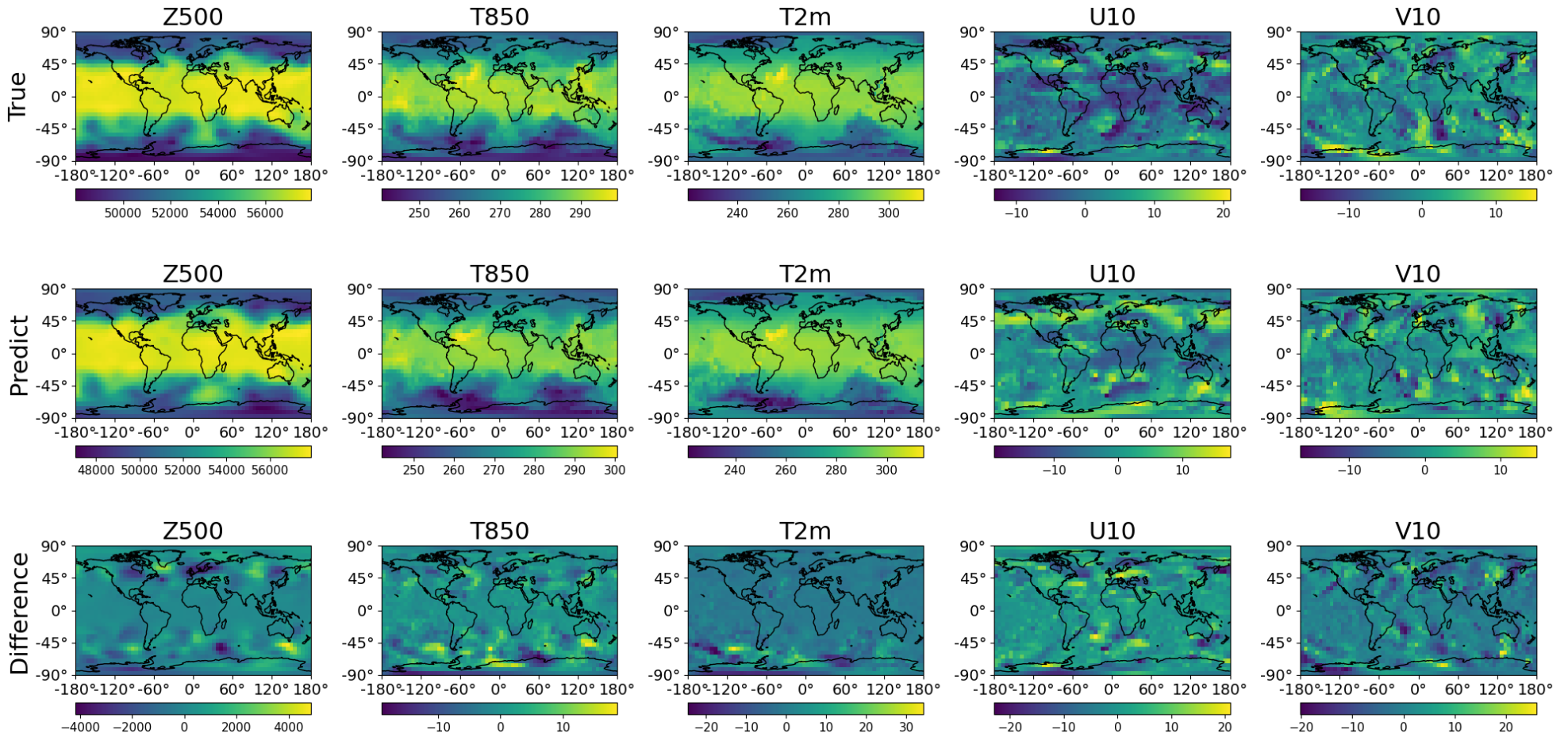}
    \caption{Visualizations of true and predicted values of all five variables at 144 hours lead time.}
    \label{fig:true_vs_pred_all_t144}
\end{figure}

\end{document}